\documentclass{esannV2}
\usepackage[dvips]{graphicx}
\usepackage[utf8]{inputenc}
\usepackage{amssymb,amsmath,array}
\usepackage[table]{xcolor}

\newcommand{\E}{\mathbb{E}}
\newcommand{\h}{\mathbf{h}}
\newcommand{\q}{\mathbf{q}}
\newcommand{\K}{\mathbf{k}}

\usepackage{graphicx}
\usepackage{caption}
\usepackage{subfig}
\usepackage{booktabs}
\usepackage{multirow}
\usepackage{xcolor}
\definecolor{darkgreen}{RGB}{0,120,10}
\definecolor{darkpurple}{RGB}{120,0,180}
\definecolor{customblue}{RGB}{0,0,200}
\definecolor{customred}{RGB}{200,0,0}

\begin{document}
\title{Stochastic Parroting in Temporal Attention -- Regulating the Diagonal Sink}

\author{Victoria Hankemeier$^1$ and Malte Schilling$^1$
%
\thanks{This research has been funded by MAN Truck \& Bus SE.}
%
\vspace{.3cm}\\
%
1- University of Münster - Faculty of Mathematics and Computer Science\\
Autonomous Intelligent Systems Group, Einsteinstr. 62, Münster - Germany\
}

\maketitle

\begin{abstract}
Spatio-temporal models analyze spatial structures and temporal dynamics, which makes them prone to information degeneration among space and time. Prior literature has demonstrated that over-squashing in causal attention or temporal convolutions creates a bias on the first tokens. To analyze whether such a bias is present in temporal attention mechanisms, we derive sensitivity bounds on the expected value of the Jacobian of a temporal attention layer. We theoretically show how off-diagonal attention scores depend on the sequence length, and that temporal attention matrices suffer a diagonal attention sink. We suggest regularization methods, and experimentally demonstrate their effectiveness.
\end{abstract}

\textbf{Note.} \textit{This paper extends our ESANN 2026 contribution by including the complete derivation of the sensitivity bounds for temporal attention and additional analysis.}

\section{Introduction}

Spatio-temporal deep learning models aim to predict how systems evolve over time and across multiple locations. They leverage historical observations as sequential input as well as using temporal information from multiple locations in order to integrate temporal dependencies as well as spatial ones. A common approach is to use Graph Neural Networks (GNNs) to capture spatial relationships where the node features are usually multivariate time series that are processed using attention or convolution-based layers to learn temporal patterns. A common challenge in spatio-temporal forecasting is to exchange information effectively between both, nearby and distant nodes with two well-known information degeneration issues: over-squashing, which is the insensitivity of a node's features to information contained at distant nodes, and over-smoothing, where node representations become increasingly similar. 
Most spatio-temporal models either mix temporal and spatial processing at every layer, i.e. Time\&Space (T\&S), or do all temporal processing first and all spatial processing afterwards, i.e. Time-then-Space (TTS). Both T\&S and TTS are affected by spatio-temporal over-squashing because their gradients still multiply across layers in the same way. Over-squashing and over-smoothing are studied extensively for spatial GNNs and Large Language Models (LLMs). However, their role in the temporal component of spatio-temporal models is not yet well understood. Previous work has focused on over-squashing in Temporal Convolutional Networks (TCNs). To address this gap, we provide a theoretical characterization of the Jacobian of Temporal Attention (TA) in which we derive why TA layers can collapse temporal information, which leads to stochastic parroting in TA and showing that this intensifies with increasing sequence length. Guided by these insights, we introduce simple architectural adjustments that significantly improve temporal information flow, mitigating over-squashing in a way complementary to existing spatial analysis of GNNs. 

\section{Previous Work on Over-squashing and Over-smoothing}

Let $\mathcal{G}_t = (\mathcal{V}, \mathcal{E}, A)$ be the weighted, directed graph at time $t$, where $\mathcal{V}$ is a set of $N$ nodes that are described by features $X_t \in \mathbb{R}^{N \times d_x}$, the edges are given in $\mathcal{E}$ and the structure is represented by the weighted adjacency matrix $A \in \mathbb{R}^{N \times N}$. The forecasting problem consists of learning a mapping $f_\theta$ from past observations $\mathcal{G}_{t-W:t}$ to an expected future sequence $\hat{\mathcal{X}}_{t:t+H} = f_\theta(\mathcal{G}_{t-W:t})$, where $W$ and $H$ denote the input and prediction horizons, respectively.

\subsection{Over-squashing and Over-smoothing in GNNs}
Di Giovanni et al. \cite{digioanni2023oversquashing} laid the groundwork for subsequent research on over-squashing by introducing sensitivity bounds on the Jacobian, which measure how sensitive the final representation of a node $v$ at layer $L$, $h^{v(L)}$, is to the initial representation of anoter node $u$, $h^{u(0)}$.
Arroyo et al. \cite{arroyo2025vanishinggradientsoversmoothingoversquashing} further showed that over-smoothing is a representational collapse to a zero fixed point, driven by the same contractive Jacobians responsible for vanishing gradients. This contractivity can lead to signals shrinking before propagating, and over-squashing occurs. For GNNs this has been solved recently (\cite{arroyo2025vanishinggradientsoversmoothingoversquashing}, 
\cite{Eliasof_AAAI2024}, \cite{Gravina_AAAI2025}, \cite{heilig2025porthamiltonian}).

\subsection{Over-squashing in Temporal Convolution GNNs}
Marisca et al. \cite{marisca2025oversquashingspatiotemporalgraphneural} is the only investigation of over-squashing in message-passing Temporal Convolution Network (MPTCNs) known to us. They show that temporal causal convolutions loose focus on recent information, while giving disproportionate weight to inputs from the distant past, creating a \textit{``primacy bias''}.

\subsection{Over-squashing in Large Language Transformers}
A similar phenomenon can be seen in causal Transformers, which only use preceding context via unidirectional attention, and auto-regressive sequence generation. In \cite{xiao2024efficient}, the emergence of an \textit{"attention sink"} in the first token is empirically demonstrated for auto-regressive Large Language Transformers. Barbero et al. \cite{barbero2024transformers} examine the Jacobian of auto-regressive decoders, showing that for a number of layers $L \to \infty$ $y_n$ is only sensitive to the first token. However, they assume attention weights independent of the input, thus focusing only on the value part, and the residual part of the attention layer's Jacobian, neglecting the key and query. In \cite{dong2021attention} the authors examine full attention and show that the residual of tokens shrink cubically per layer with a contraction factor proportional to $\frac{4 \beta}{\sqrt{d_{qk}}}$, where $\beta$ is the upper bound of the norm of weight matrices. Thus, smaller values of $beta$ and larger key-query dimensions $d_{qk}$ accelerate convergence toward a rank-1 representation. As a consequence residual (skip) connections and and appropriate choices of network dimensionalities are important to avoid rank collapse. Besides, the largest eigenvalue of the softmax attention matrix differs significantly in magnitude from the second largest, and this gap widens as the context length increases \cite{saada2025mind}. As a result, token representations collapse to a single direction as the context length increases. Removing the dominant rank-1 component of the attention matrix has become the standard solution. Another study on token similarity was made in \cite{wu2024role} with focus on attention masks and layer normalization. Pure self-attention with causal or sparse masks still converges exponentially to rank-1 while layer normalization may break the collapse.

\section{Over-squashing in Temporal Attention GNNs} \label{seq: 3}
The previously shown \textit{primacy bias} does not appear in TA layers with a full attention matrix. As we will derive in the following, the temporal attention has a \textit{diagonal attention sink}, which for very large sequence lengths results in self-copying behavior, i.e. \textit{stochastic parroting}. However, as shown in \textit{SparseBERT} \cite{shi2021sparsebert} bidirectional Transformers, with a full attention matrix and residual, remain universally expressible even when the diagonal is completely removed, suggesting that the diagonal provides the least important information.

\subsection{Deriving Sensitivity Bounds of Temporal Attention}
We examine the influence of $\mathbf{x}_j \in \mathbb{R}^{d_{\text{x}}}$ for an input $\mathbf{x}_i \in \mathbb{R}^{d_{\text{x}}}$ on the hidden state $\mathbf{h}_i \in \mathbb{R}^{d_v}$ after one layer by deriving the Jacobian. 
The index $i$ denotes the source time step, and $j$ the target time step.
The hidden state is a weighted sum of the value vectors, where the index $m$ is the summation index that iterates over all source time steps in the sequence, from $m=1$ to $T$.
With the weight matrices $W^Q \in \mathbb{R}^{d_k \times d_{\text{model}}}$, $W^K \in \mathbb{R}^{d_k \times d_{\text{model}}}$, and $W^V \in \mathbb{R}^{d_v \times d_{\text{model}}}$, we get the query $\mathbf{q}_i = W^Q \mathbf{x}_i$, key $\mathbf{k}_j = W^K \mathbf{x}_j$ and value $\mathbf{v}_j = W^V \mathbf{x}_j$. The scalar attention score of time steps $i$ and $j$ is $e_{ij} = (\mathbf{q}_i^{\!\top} \mathbf{k}_j) / \sqrt{d_k}$. The softmax attention is $\alpha_{ij} = \exp(e_{ij}) \big/ \sum_{m=1}^T \exp(e_{im})$, and the hidden state is $ \mathbf{h}_i = \sum_{m=1}^T \alpha_{im} \mathbf{v}_m$, with $\sum_{m=1}^T \alpha_{im}=1$.
\begin{equation}
    \frac{\partial \mathbf{h}_i}{\partial \mathbf{x}_j} = \frac{\partial}{\partial \mathbf{x}_j} \left( \sum_{m=1}^T \alpha_{im} \mathbf{v}_m \right) = \sum_{m=1}^T \left[
        \underbrace{\alpha_{im} \frac{\partial \mathbf{v}_m}{\partial \mathbf{x}_j}}_{\text{\texttt{Value Path}}}
        + 
        \underbrace{\mathbf{v}_m \frac{\partial \alpha_{im}}{\partial \mathbf{x}_j}}_{\text{\texttt{Weight Path}}}
      \right]
\end{equation}
The Jacobian of the \texttt{Value Path} is:
\begin{equation}
    \sum_{m=1}^T \alpha_{im} \frac{\partial \mathbf{v}_m}{\partial \mathbf{x}_j}
    = \alpha_{ij} W^V
\end{equation}
The Jacobian of the \texttt{Weight Path} consists of the softmax and the score gradient:
\begin{equation}
    \frac{\partial \alpha_{im}}{\partial e_{ik}} = \alpha_{im} (\delta_{mk} - \alpha_{ik}) \qquad
    \frac{\partial e_{ik}}{\partial \mathbf{x}_j}= \frac{1}{\sqrt{d_k}} \left( \delta_{ij} \mathbf{k}_k^{\!\top} W^Q + \delta_{kj} \mathbf{q}_i^{\!\top} W^K \right)
\end{equation}
The overall Jacobian without residual is thus:
\begin{equation}
    \frac{\partial \mathbf{h}_i}{\partial \mathbf{x}_j} =
    \underbrace{\alpha_{ij} W^V}_{\text{\texttt{Value Path}}}
    +
    \sum_{m=1}^T 
    \left[
        \mathbf{v}_m
        \left(
        \sum_{k=1}^T
        \underbrace{\alpha_{im}(\delta_{mk} - \alpha_{ik})}_{\text{\texttt{Softmax Gradient}}}
        \cdot
        \underbrace{
        \frac{1}{\sqrt{d_k}} \left(
        \underbrace{
        \delta_{ij} \mathbf{k}_k^T W^Q }_{\text{\texttt{Query Path}}}
        +
        \underbrace{
        \delta_{kj} \mathbf{q}_i^T W^K }_{\text{\texttt{Key Path}}} \right)
        }_{\text{\texttt{Score Gradient}}}
        \right)
    \right]
\end{equation}

\begin{equation*}
    \frac{\partial \mathbf{h}_i}{\partial \mathbf{x}_j} =
    \underbrace{\alpha_{ij} W^V}_{\textcolor[HTML]{008E96}{\text{\texttt{Value Path}}}}
    +
    \sum_{m=1}^T 
    \left[
        \mathbf{v}_m
        \left(
        \sum_{k=1}^T
        \underbrace{\alpha_{im}(\delta_{mk} - \alpha_{ik})}_{\textcolor[HTML]{008E96}{\text{\texttt{Softmax Gradient}}}}
        \cdot
        \underbrace{
        \frac{1}{\sqrt{d_k}} \left(
        \underbrace{
        \delta_{ij} \mathbf{k}_k^T W^Q }_{\textcolor[HTML]{008E96}{\text{\texttt{Query Path}}}}
        +
        \underbrace{
        \delta_{kj} \mathbf{q}_i^T W^K }_{\textcolor[HTML]{008E96}{\text{\texttt{Key Path}}}} \right)
        }_{\textcolor[HTML]{008E96}{\text{\texttt{Score Gradient}}}}
        \right)
    \right]
\end{equation*}

We derive a uniform expectation of the Jacobian norm over the sequence length $j \in \{1,\ldots,T\}$ for the \texttt{Value Path}, \texttt{Key Path}, and \texttt{Query Path} separately:
\begin{align}
    &\E\big[\|J_{ij}^{(\mathrm{value})}\|\big]
    = \frac{1}{T}\sum_{j=1}^{T} \alpha_{ij}\,\|W^V\|
    = \frac{\|W^V\|}{T}
    \\[-0.6ex]
    &\E\!\left[\big\|J_{ij}^{(\text{key})}\big\|\right]
    \;=\; \frac{1}{T\sqrt{d_k}} \sum_{j=1}^T \alpha_{ij}
      \left( \|\mathbf{v}_j-\h_i\|\;\cdot\;\|(W^K)^{\!\top}\q_i\| \right)
      \;\le\; \frac{C_K}{T\sqrt{d_k}}
    \\[-0.6ex]
    &\E\!\left[\big\|J_{ij}^{(\text{query})}\big\|\right]
    \;\le\; \frac{1}{T\sqrt{d_k}} \sum_{k=1}^T \alpha_{ik}
      \left( \|\mathbf{v}_k - \h_i\| \cdot \|(W^Q)^{\!\top} \K_k\| \right)
      \;\le\; \frac{C_Q}{T\sqrt{d_k}}
\end{align}
with $C_K := \max_j \| \mathbf{v}_j-\h_i \|\;\|(W^K)^{\!\top}\q_i\|,
\quad
C_Q := \max_k \| \mathbf{v}_k-\h_i \|\;\|(W^Q)^{\!\top}\K_k\|$.
We distinguish the diagonal case ($i=j$), corresponding to local self-updates,
and the off-diagonal case ($i\neq j$), corresponding to non-local interactions.
The query vector $\mathbf{q}_i$ is defined as $\mathbf{q}_i = W^Q \mathbf{x}_i$. Thus, the derivative of $\mathbf{q}_i$ with respect to $\mathbf{x}_j$ is zero if $i \neq j$, because $\mathbf{q}_i$ is independent of $\mathbf{x}_j$. We define bounds on the expected value of the Jacobian of the attention layer with a residual connection, using the expected norm for off-diagonal terms ($i \neq j$) and the norm for the diagonal term ($i = j$).\footnote{The corresponding code is published here: https://github.com/vicky-hnk/spatio-temp-parroting.}
\begin{equation}
\label{eq:jacobian-bounds}
\E\!\left[\|\delta_{ij} I + J_{ij}^{(\mathrm{total})}\|\right]
\le
\begin{cases}
\dfrac{1}{T}\!\left(\|W^V\| + \dfrac{C_K}{\sqrt{d_k}}\right), 
& i \neq j, \\[0.6ex]
\|I\|
      + \E\!\left[\alpha_{ii}\right]\!\left(\|W^V\| 
        + \dfrac{C_K}{\sqrt{d_k}} 
        + \dfrac{C_Q}{\sqrt{d_k}}\right),
& i = j .
\end{cases}
\end{equation}

\subsection{Analysis and Regularization of Diagonal Attention Sinks}
The $1/T$-term in the upper part of Eq. \ref{eq:jacobian-bounds} shows that the non-local signal strength ($i \neq j$) decays as the sequences are longer ($O(1/T)$). The lower part of Eq. \ref{eq:jacobian-bounds} for the diagonal ($i = j$) has a larger bound since it additionally has the query term's bound $C_Q / \sqrt{d_k}$. With a residual path, self-information remains stable regardless of sequence length ($O(1)$). 
$\alpha_{ii}$ often exceeds $\alpha_{ij}$ as positional encoding (PE) favors information near the diagonal. While emphasizing the diagonal makes sense given the importance of nearby time steps, the extreme case of self-copying does not.
To reduce the diagonal attention sink, we evaluated three different regularization methods: a diagonal mask with diagonal entries $-\infty$ and others zero, similar to the approach in \textit{SparseBert} \cite{shi2021sparsebert}, standard dropout on the diagonal of the attention matrix, and a negative scalar penalty added to the raw attention scores.

\section{Empirical Evaluation} \label{seq: 4}
We evaluate our model on the METR-LA traffic dataset, as described in  \cite{li2018dcrnn}. The forecasting horizon is 12 steps, using the previous 12 steps as input. Optimization uses AdamW with cosine scheduling, 5 warm-up epochs, 150 epochs in total, and an initial learning rate $0.001$. We stack a temporal softmax attention block followed by a Graph Convolution Network with learned forward and backward adjacency matrix (T\&S). The number of TA heads is $8$. We reduce the spatial influence by using a single-layer GCN with about four times fewer parameters than the TA module. The optimal diagonal penalty and dropout were $-0.1$ and $0.2$, respectively. Since relative PE introduces another bias towards the diagonal, we used absolute PE (tAPE from \cite{foumani2024position}).

\subsection{Results}
\begin{figure}[h!]
\centering
\subfloat[\centering No residual\label{fig:res}]{
  \includegraphics[scale=0.15]{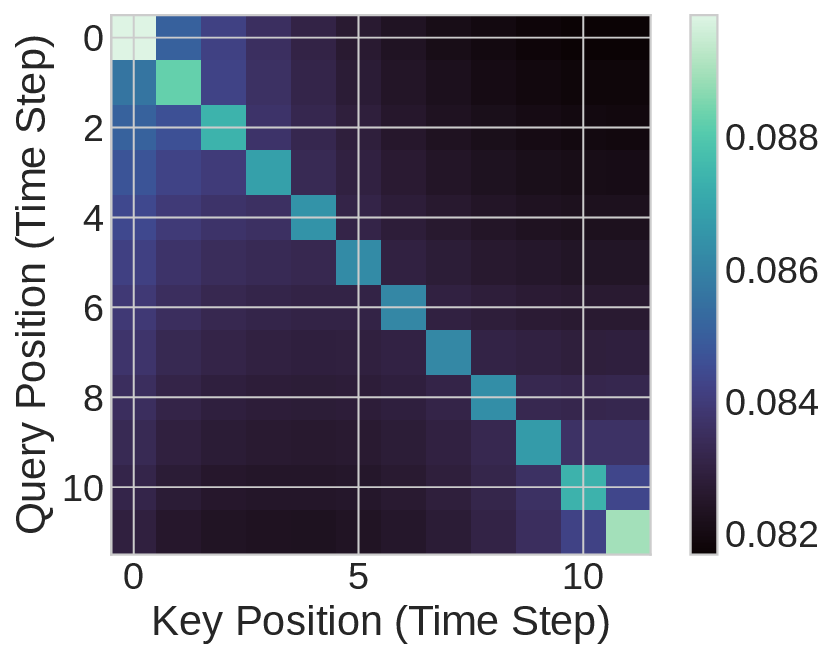}
}
\subfloat[\centering No regularization\label{fig:wo}]{
  \includegraphics[scale=0.15]{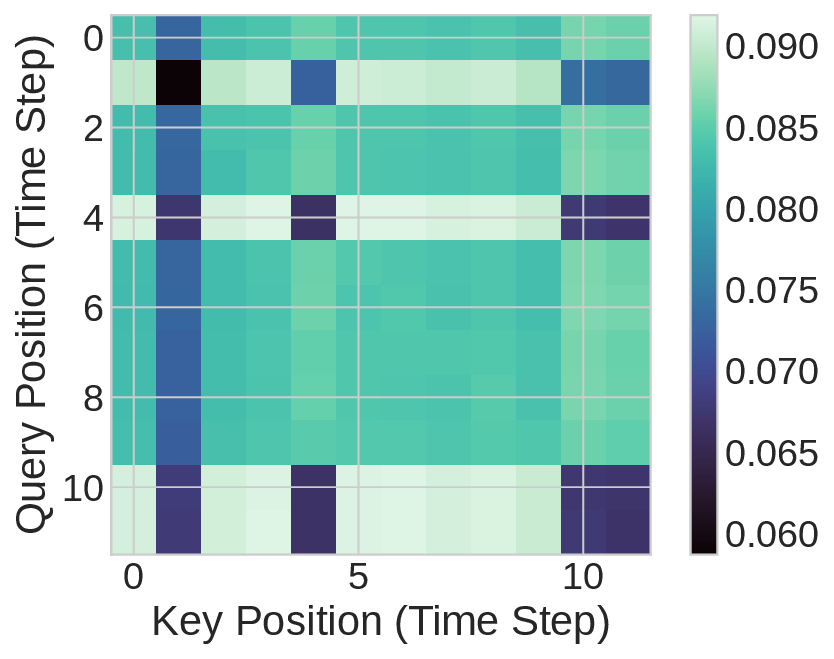}
}
\subfloat[\centering Mask\label{fig:mask}]{
  \includegraphics[scale=0.15]{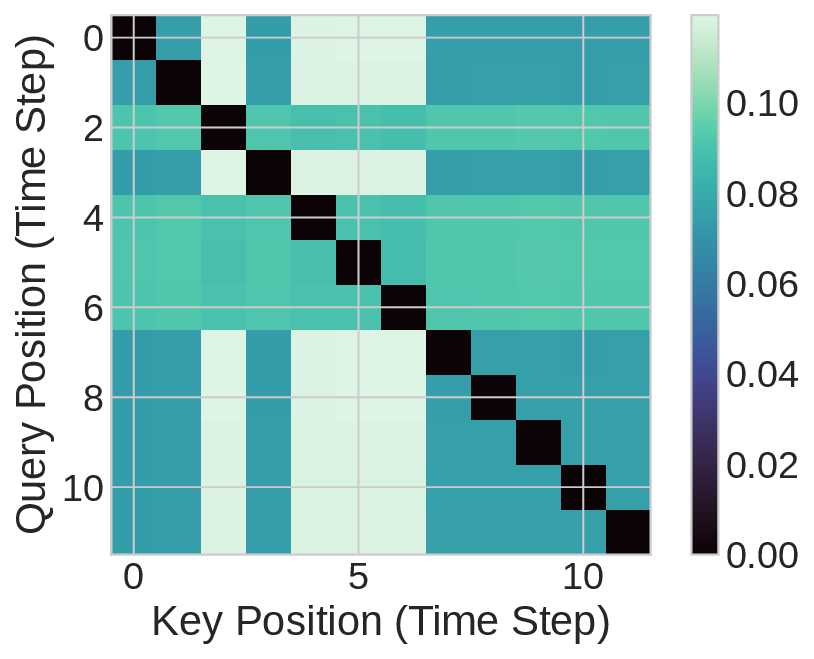}
}
\subfloat[\centering Penalty\label{fig:pen}]{
  \includegraphics[scale=0.15]{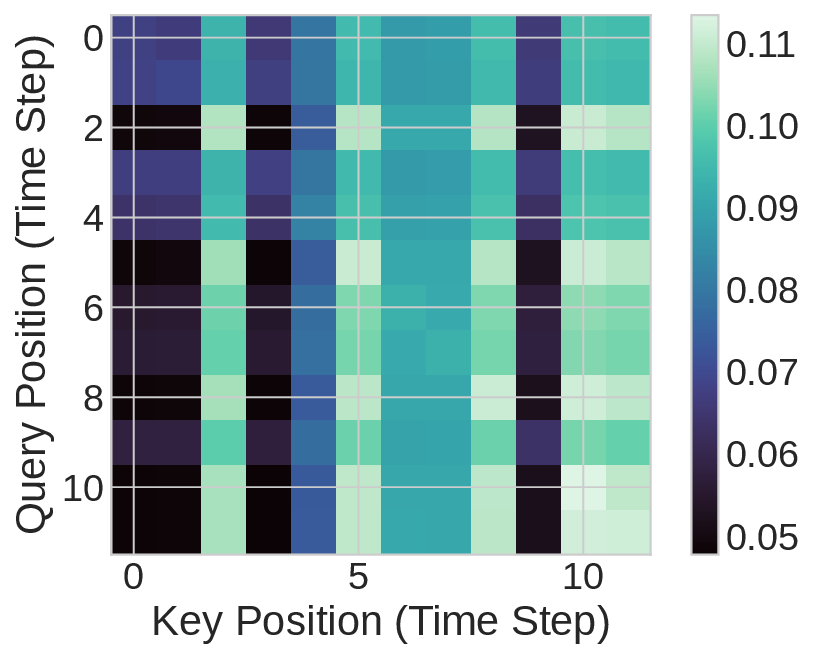}
}
\subfloat[\centering Diagonal dropout\label{fig:drop}]{
  \includegraphics[scale=0.15]{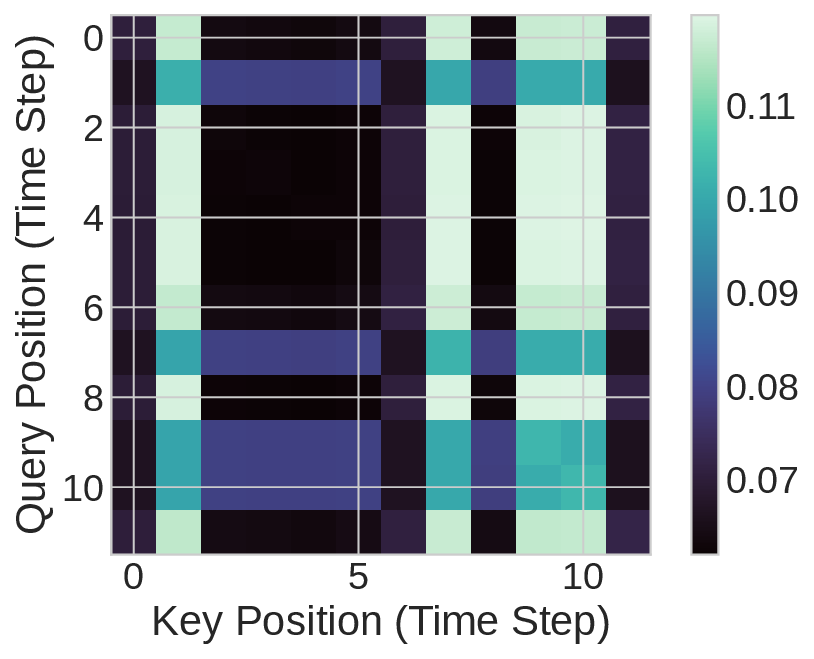}
}
\caption{Attention matrices under different regularizations.}
\label{fig:all}
\end{figure}

Table  \ref{tab:reg_metrics_horizons} shows the results over four runs with different seeds for a) without residual and without regularization, b) with residual and without regularization, c) with residual and with diagonal mask, d) with residual and with diagonal dropout, and e) with residual and diagonal penalty. The values are the Mean Absolute Error (MAE), Root Mean Squared Error (RMSE) and Mean Absolute Percentage Error (MAPE) for the forecasted steps $3$, $6$ and $12$. The baseline model without residuals shows substantially higher error than models with residual connections. Using a diagonal mask (c) gives similar results as the variant with residual but without regularization, while dropout and penalty show a significant improvement of around $2.5$ percent. Fig. \ref{fig:all} shows the heat maps of final attention matrices after the last epoch. In (a) we see that the attention scores on the diagonal are very large and get smaller with distance if no residual term is used. Fig. \ref{fig:all} (b) shows the attention matrix with residual but without regularization. It is diffuse without clear temporal patterns, as the residual already provides a stable identity mapping. By enforcing regularization on the diagonal elements of the weight matrix, we structurally incentivize the model to allocate learning capacity towards the off-diagonal entries. However, our experiments suggest that the partial control of the diagonal using dropout or penalty is superior to a full diagonal mask. With a full diagonal mask (c) we see similar results as without regularization (b). By masking the diagonal, we suppress the \texttt{Query Path}, making the attention less expressive (see Eq. \ref{eq:jacobian-bounds}). In (d) and (e) we see clear temporal patterns with specific keys attending to specific queries. 

\begin{table}[t!b]
\centering
\renewcommand{\arraystretch}{1.1}
\captionsetup{format=hang,skip=4pt} 

\caption{Performance across forecasting horizons (3, 6, 12). MAE, RMSE, and MAPE reported with mean $\pm$ standard deviation.}

\resizebox{\columnwidth}{!}{
\begin{tabular}{l|ccc|ccc|ccc}
\hline
\multirow{2}{*}{\textbf{Model}} &
\multicolumn{3}{c|}{\textbf{MAE}} &
\multicolumn{3}{c|}{\textbf{RMSE}} &
\multicolumn{3}{c}{\textbf{MAPE}} \\
 & \textbf{3} & \textbf{6} & \textbf{12}
 & \textbf{3} & \textbf{6} & \textbf{12}
 & \textbf{3} & \textbf{6} & \textbf{12} \\
\hline

(a) No residual &
$3.165{\scriptstyle\pm0.028}$ & $3.671{\scriptstyle\pm0.064}$ & $4.627{\scriptstyle\pm0.113}$ &
$6.190{\scriptstyle\pm0.050}$ & $7.414{\scriptstyle\pm0.144}$ & $9.315{\scriptstyle\pm0.227}$ &
$8.705{\scriptstyle\pm0.268}$ & $10.286{\scriptstyle\pm0.202}$ & $13.406{\scriptstyle\pm0.185}$ \\

(b) No regularization &
$2.966{\scriptstyle\pm0.022}$ & $3.443{\scriptstyle\pm0.006}$ & $3.989{\scriptstyle\pm0.019}$ &
$5.881{\scriptstyle\pm0.045}$ & $7.028{\scriptstyle\pm0.054}$ & $8.418{\scriptstyle\pm0.053}$ &
$7.836{\scriptstyle\pm0.041}$ & $9.670{\scriptstyle\pm0.082}$ & $11.708{\scriptstyle\pm0.158}$ \\

(c) Full diagonal mask &
$2.969{\scriptstyle\pm0.019}$ & $3.447{\scriptstyle\pm0.022}$ & $3.997{\scriptstyle\pm0.026}$ &
$5.889{\scriptstyle\pm0.015}$ & $7.000{\scriptstyle\pm0.017}$ & $8.431{\scriptstyle\pm0.024}$ &
$\mathbf{7.830{\scriptstyle\pm0.060}}$ & $9.666{\scriptstyle\pm0.100}$ & $11.734{\scriptstyle\pm0.083}$ \\

(d) Diagonal dropout &
$\mathbf{2.959{\scriptstyle\pm0.001}}$ & $\mathbf{3.415{\scriptstyle\pm0.010}}$ & $\mathbf{3.940{\scriptstyle\pm0.011}}$ &
$\mathbf{5.845{\scriptstyle\pm0.029}}$ & $7.006{\scriptstyle\pm0.051}$ & $\mathbf{8.338{\scriptstyle\pm0.041}}$ &
$7.833{\scriptstyle\pm0.003}$ & $\mathbf{9.616{\scriptstyle\pm0.037}}$ & $\mathbf{11.570{\scriptstyle\pm0.047}}$ \\

(e) Diagonal penalty &
$2.969{\scriptstyle\pm0.027}$ & $3.436{\scriptstyle\pm0.037}$ & $3.964{\scriptstyle\pm0.041}$ &
$5.859{\scriptstyle\pm0.034}$ & $\mathbf{6.995{\scriptstyle\pm0.054}}$ & $8.344{\scriptstyle\pm0.105}$ &
$7.842{\scriptstyle\pm0.070}$ & $9.655{\scriptstyle\pm0.120}$ & $11.623{\scriptstyle\pm0.157}$ \\
\hline

\end{tabular}
}
\label{tab:reg_metrics_horizons}
\end{table}

\section{Conclusion}
Recent work on over-squashing in spatio-temporal models has focused on Temporal Convolution \cite{marisca2025oversquashingspatiotemporalgraphneural}. We advance the understanding of information degeneration in spatio-temporal models by deriving sensitivity bounds on the Jacobian of temporal attention layers. We prove the existence of diagonal attention sinks, induced by the residual and the weight matrix of the query. While full diagonal masks suppress the query part of the softmax attention, resulting in attention emphasizing specific keys for the whole sequence, diagonal penalty and diagonal dropout significantly improve the performance of temporal attention layers.


\begin{footnotesize}





\end{footnotesize}


\appendix
\section{Full derivation of Jacobian and sensitivity bounds}
In Section \ref{seq: 3} we analyzed the sensitivity bounds of temporal attention. In the following we first present the complete derivation of the Jacobian in Section \ref{seq: jacob} and proceed with the derivation of the sensitivity bounds in Section \ref{seq: bounds}.

\subsection{Derivation of the Attention Layer Jacobian} \label{seq: jacob}
\begin{equation}
    \frac{\partial \mathbf{h}_i}{\partial \mathbf{x}_j} = \frac{\partial}{\partial \mathbf{x}_j} \left( \sum_{m=1}^T \alpha_{im} \mathbf{v}_m \right) = \sum_{m=1}^T \left[
        \underbrace{\alpha_{im} \frac{\partial \mathbf{v}_m}{\partial \mathbf{x}_j}}_{\text{\texttt{Value Path}}}
        + 
        \underbrace{\mathbf{v}_m \frac{\partial \alpha_{im}}{\partial \mathbf{x}_j}}_{\text{\texttt{Weight Path}}}
      \right]
\end{equation}
We begin with the value path:
\begin{equation}
    J_{ij}^{value} = \alpha_{ij} W^V
\end{equation}
The Jacobian of the weight path can be derived as follows:
\begin{align}
    J_{ij}^{weight} &= \sum_{m = 1}^{T} v_m \frac{\partial \alpha_{im}}{\partial x_j} \\
    &=  \sum_{m = 1}^{T} v_m  \sum_{k = 1}^{T} \frac{\partial\alpha_{im}}{\partial e_{ik}} \frac{\partial e_{ik}}{\partial x_j}
\end{align}
We derive the chain rule factors separately, first the softmax gradient:
\begin{equation}
    \frac{\partial\alpha_{im}}{\partial e_{ik}} = \alpha_{im} (\delta_{mk} - \alpha_{ik})
\end{equation}
The score gradient is given as:
\begin{equation}
    \frac{\partial e_{ik}}{\partial x_j} = \frac{1}{\sqrt{d_k}} \left(
        \underbrace{
        \delta_{ij} \mathbf{k}_k^T W^Q }_{\text{\texttt{Query Path}}}
        +
        \underbrace{
        \delta_{kj} \mathbf{q}_i^T W^K }_{\text{\texttt{Key Path}}} \right)
\end{equation}

The overall Jacobian is given by:
\begin{equation}
    \frac{\partial \mathbf{h}_i}{\partial \mathbf{x}_j} =
    \underbrace{\alpha_{ij} W^V}_{\text{\texttt{Value Path}}}
    +
    \sum_{m=1}^T 
    \left[
        \mathbf{v}_m
        \left(
        \sum_{k=1}^T
        \underbrace{\alpha_{im}(\delta_{mk} - \alpha_{ik})}_{\text{\texttt{Softmax Gradient}}}
        \cdot
        \underbrace{
        \frac{1}{\sqrt{d_k}} \left(
        \underbrace{
        \delta_{ij} \mathbf{k}_k^T W^Q }_{\text{\texttt{Query Path}}}
        +
        \underbrace{
        \delta_{kj} \mathbf{q}_i^T W^K }_{\text{\texttt{Key Path}}} \right)
        }_{\text{\texttt{Score Gradient}}}
        \right)
    \right]
\end{equation}

\subsection{Deriving the Sensitivity bounds} \label{seq: bounds}

To determine the Jacobian bounds, we examine the query, value, and key terms separately:
\begin{align}
J_{ij}^{(\text{query})}
&= \sum_{m=1}^T \mathbf{v}_m
   \sum_{k=1}^T
   \alpha_{im}(\delta_{mk}-\alpha_{ik})
   \frac{1}{\sqrt{d_k}}
   \delta_{ij}(W^Q)^\top \mathbf{k}_k
\\
&= \frac{\delta_{ij}}{\sqrt{d_k}}
   \sum_{m=1}^T
   \alpha_{im}\mathbf{v}_m
   \Bigg(
      (W^Q)^\top \mathbf{k}_m
      - \sum_{k=1}^T \alpha_{ik}(W^Q)^\top \mathbf{k}_k
   \Bigg)
\\
&= \frac{\delta_{ij}}{\sqrt{d_k}}
   \sum_{m=1}^T
   \alpha_{im}\mathbf{v}_m
   \otimes
   (W^Q)^\top
   \Bigg(
      \mathbf{k}_m - \sum_{k=1}^T \alpha_{ik}\mathbf{k}_k
   \Bigg)
\\
&= \frac{\delta_{ij}}{\sqrt{d_k}}
   \sum_{m=1}^T
   \alpha_{im}\mathbf{v}_m
   \otimes
   (W^Q)^\top
   (\mathbf{k}_m-\bar{\mathbf{k}}_i),
\end{align}
with $\bar{\mathbf{k}}_i := \sum_{k=1}^T \alpha_{ik}\mathbf{k}_k$.\\
\\
The Jacobian of the value term is given by:
\begin{align}
J_{ij}^{(\text{value})} = \alpha_{ij} W^V .
\end{align}
and of the key term by:
\begin{align}
J_{ij}^{(\text{key})}
&= \sum_{m=1}^T \mathbf{v}_m
   \sum_{k=1}^T
   \alpha_{im}(\delta_{mk}-\alpha_{ik})
   \frac{1}{\sqrt{d_k}}
   \delta_{kj}(W^K)^\top \mathbf{q}_i
\\
&= \frac{\alpha_{ij}}{\sqrt{d_k}}
   (\mathbf{v}_j-\mathbf{h}_i)
   \otimes (W^K)^\top \mathbf{q}_i ,
\end{align}

with $\mathbf{h}_i := \sum_{m=1}^T \alpha_{im}\mathbf{v}_m.$\\
\\
We derive a uniform expectation of the Jacobian norm over the sequence length $j \in \{1,\ldots,T\}$:

\begin{align}
\mathbb{E}\!\left[\big\|J_{ij}^{(\text{query})}\big\|\right]
&= \frac{1}{T}
   \left\|
   \frac{1}{\sqrt{d_k}}
   \sum_{m=1}^T
   \alpha_{im}\mathbf{v}_m
   \otimes
   (W^Q)^\top(\mathbf{k}_m-\bar{\mathbf{k}}_i)
   \right\|
\\
&\le \frac{1}{T\sqrt{d_k}}
   \sum_{m=1}^T
   \alpha_{im}
   \|\mathbf{v}_m-\mathbf{h}_i\|
   \cdot
   \|(W^Q)^\top \mathbf{K}_m\|
\\
&\le \frac{1}{T\sqrt{d_k}}
   \max_m
   \Big(
      \|\mathbf{v}_m-\mathbf{h}_i\|
      \|(W^Q)^\top \mathbf{K}_m\|
   \Big)
   \sum_{m=1}^T \alpha_{im}
\\
&= \frac{C_Q}{T\sqrt{d_k}},
\end{align}
with $C_Q := \max_m
\|\mathbf{v}_m-\mathbf{h}_i\|
\;\|(W^Q)^\top \mathbf{K}_m\|.$\\
\\
The value term can be bound with:
\begin{equation}
\E\big[\|J_{ij}^{(\mathrm{value})}\|\big]
    = \frac{1}{T}\sum_{j=1}^{T} \alpha_{ij}\,\|W^V\|
    = \frac{\|W^V\|}{T}
\end{equation}
and the key term with:
\begin{align}
\mathbb{E}\!\left[\|J_{ij}^{(\text{key})}\|\right]
&= \frac{1}{T}
   \sum_{j=1}^T
   \left\|
   \frac{\alpha_{ij}}{\sqrt{d_k}}
   (\mathbf{v}_j-\mathbf{h}_i)
   \otimes (W^K)^\top \mathbf{q}_i
   \right\|
\\
&\le \frac{1}{T\sqrt{d_k}}
   \sum_{j=1}^T
   \alpha_{ij}
   \|\mathbf{v}_j-\mathbf{h}_i\|
   \cdot
   \|(W^K)^\top \mathbf{q}_i\|
\\
&\le \frac{C_K}{T\sqrt{d_k}},
\end{align}
with $C_K := \max_j \| \mathbf{v}_j-\h_i \|\;\|(W^K)^{\!\top}\q_i\|$\\

\section{Evaluation for Long Horizon Forecasting}
As a robustness check, we extended the evaluation to long-horizon forecasting, predicting 36 future steps (i.e., 3 hours). We used the same experimental set-up and evaluation methods as in Section \ref{seq: 4} for the short-term forecasting of 12 time steps.\\
\\

\begin{figure}[h!]
\centering
\subfloat[\centering No residual\label{fig:res36}]{
  \includegraphics[scale=0.4]{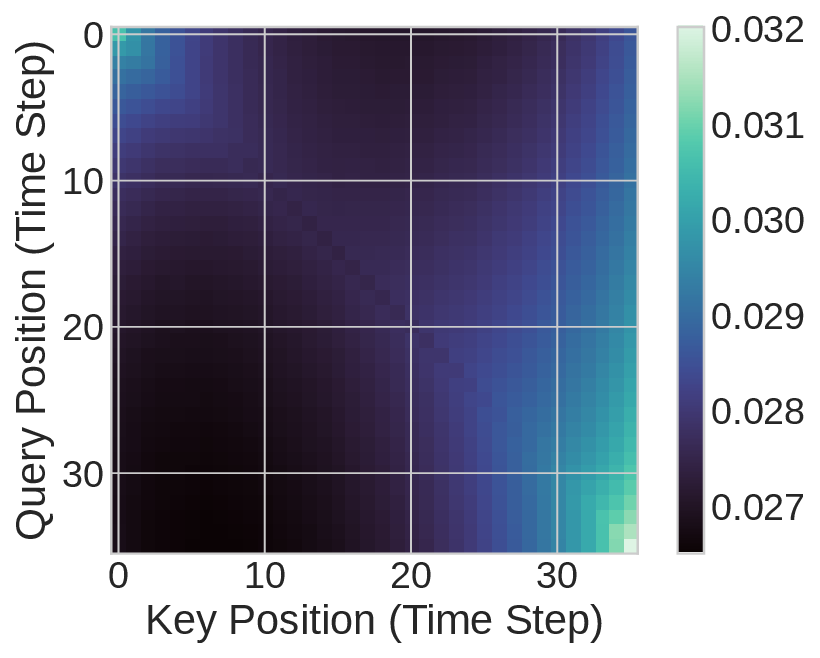}
}\hfill
\subfloat[\centering No regularization\label{fig:wo36}]{
  \includegraphics[scale=0.4]{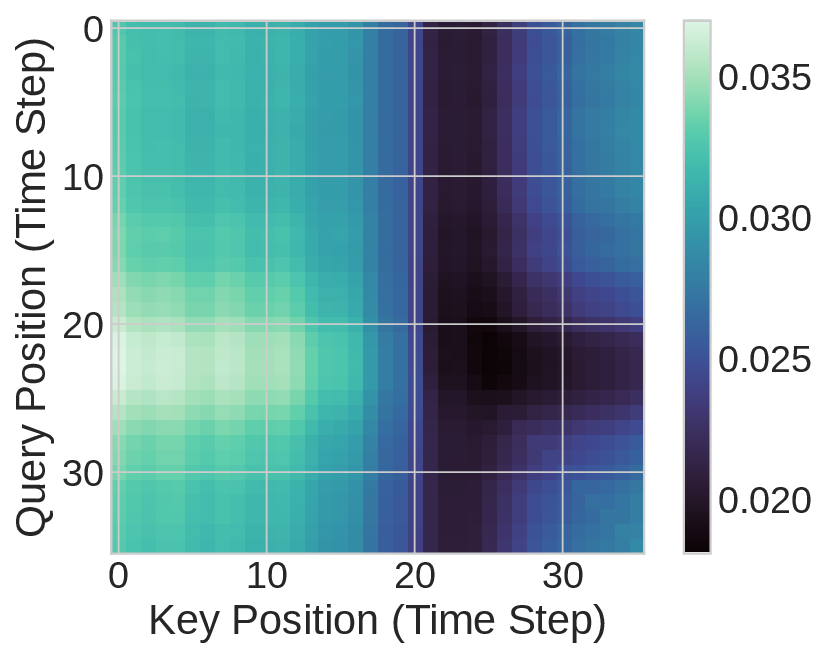}
}

\vspace{0.3cm}

\subfloat[\centering Mask\label{fig:mask36}]{
  \includegraphics[scale=0.4]{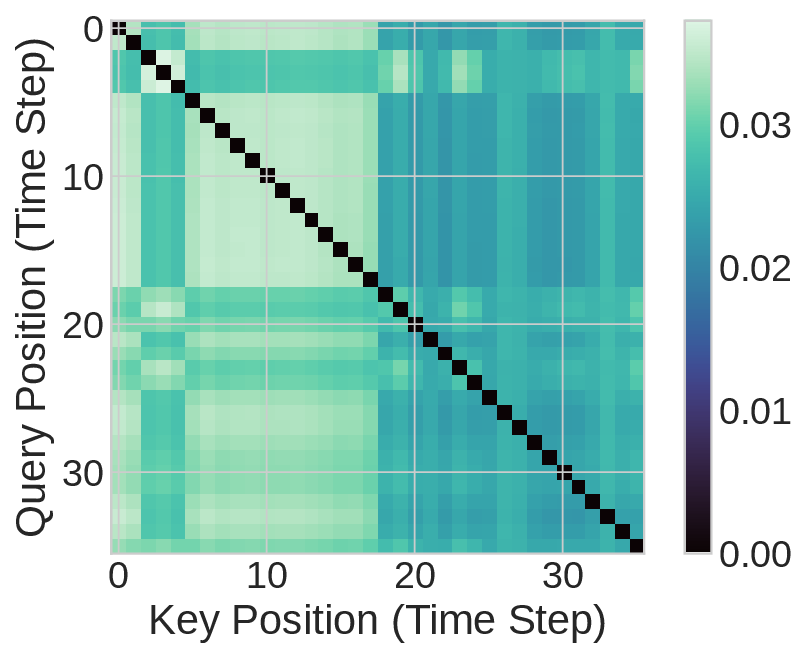}
}\hfill
\subfloat[\centering Penalty\label{fig:pen36}]{
  \includegraphics[scale=0.4]{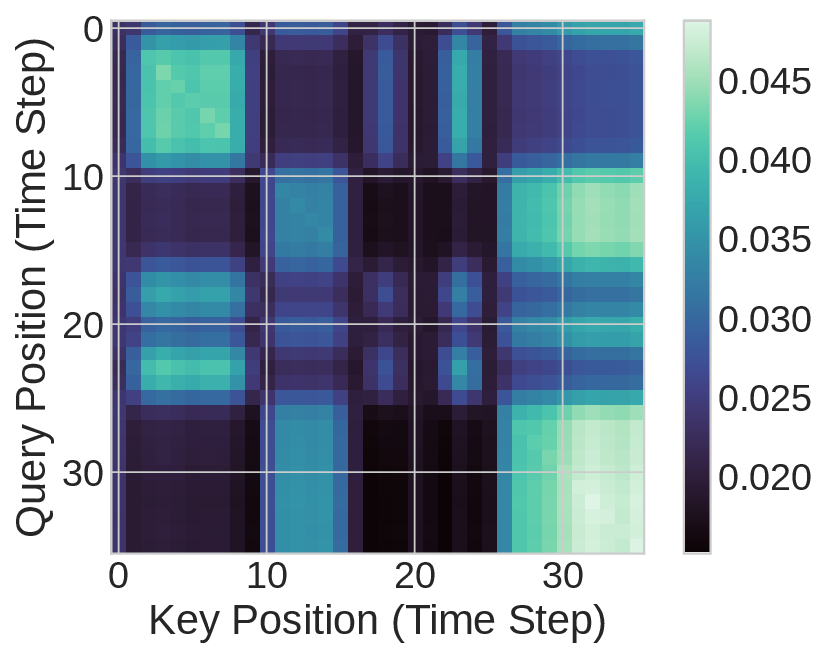}
}

\vspace{0.3cm}

\subfloat[\centering Diagonal dropout\label{fig:drop36}]{
  \includegraphics[scale=0.4]{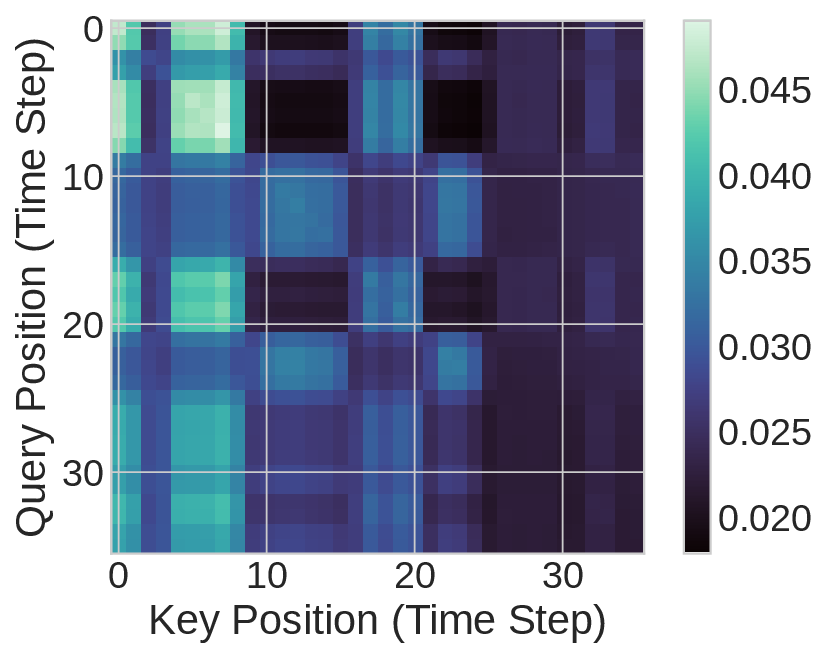}
}

\caption{Attention matrices under different regularizations for long horizons.}
\label{fig:all36}
\end{figure}

Table \ref{tab:long_horizons} reports the results obtained by training the model with a sequence length of 36 time steps and forecasting 36 future steps. The Mean Absolute Error (MAE), Root Mean Squared Error (RMSE), and Mean Absolute Percentage Error (MAPE) are reported for the 24th and 36th forecasting steps. In Section \ref{seq: 4} Table \ref{tab:reg_metrics_horizons} we can see that the results with residual are better than without. For longer horizons the difference between the results without a residual (a) and with residual (b-e) becomes even more significant. This aligns well with the rank collapse in width shown in \cite{saada2025mind}, as well as with our previous findings. The larger $T$ is, the smaller off-diagonal entries become. \\
\\
This can also be seen in Figure \ref{fig:res36}. Overall, the diagonal penalty gives best results. For longer horizons, we have more heterogeneous patterns, and the diagonal might provide an important anchor. Using a penalty enables the model to use the diagonal entries when beneficial, which is not possible with dropout or masking.

\begin{table}[t!b]
\centering
\footnotesize
\renewcommand{\arraystretch}{1.05}
\captionsetup{format=hang,skip=4pt}
\caption{Performance across forecasting horizons (24, 36). MAE, RMSE, and MAPE reported as mean $\pm$ standard deviation.}
\resizebox{\columnwidth}{!}{
\begin{tabular}{lccc ccc ccc}
\toprule
\multirow{2}{*}{\textbf{Model}} &
\multicolumn{2}{c}{\textbf{MAE}} &
\multicolumn{2}{c}{\textbf{RMSE}} &
\multicolumn{2}{c}{\textbf{MAPE}} \\
 & \textbf{24} & \textbf{36}
 & \textbf{24} & \textbf{36}
 & \textbf{24} & \textbf{36} \\
\midrule

(a) No residual
 & $5.6906$ & $6.7430$
 & $10.9244$ & $12.4455$
 & $16.1033$ & $19.5359$ \\

(b) No regularization
& $4.4438$ & $4.7241$
& $\mathbf{9.0802}$ & $9.7268$
& $13.0853$ & $14.0347$ \\

(c) Full diagonal mask
& $4.4761$ & $4.7784$
& $9.2640$ & $9.7898$
& $13.1699$ & $\mathbf{14.1011}$ \\

(d) Diagonal dropout
& $4.4788$ & $4.7729$
& $9.0856$ & $9.8035$
& $13.3252$ & $14.3873$ \\

(e) Diagonal penalty
& $\mathbf{4.4058}$ & $\mathbf{4.6856}$
& $9.1346$ & $\mathbf{9.6846}$
& $\mathbf{13.1030}$ & $14.1929$ \\

\bottomrule
\end{tabular}
}
\label{tab:long_horizons}
\end{table}

\begin{align}
\frac{\partial \rho}{\partial t} + \frac{\partial (\rho v)}{\partial x} &= 0, \\
\frac{\partial v}{\partial t} + v \frac{\partial v}{\partial x} &=
- \frac{1}{\rho} \frac{\partial P(\rho)}{\partial x}
+ \frac{V_{\mathrm{eq}}(\rho) - v}{\tau}
+ \nu \frac{\partial^2 v}{\partial x^2}.
\end{align}

\end{document}